\documentclass[pdflatex,sn-mathphys-num]{sn-jnl}
\usepackage{graphicx}%
\usepackage{multirow}%
\usepackage{amsmath,amssymb,amsfonts}%
\usepackage{amsthm}%
\usepackage{mathrsfs}%
\usepackage[title]{appendix}%
\usepackage{xcolor}%
\usepackage{textcomp}%
\usepackage{booktabs}%
\usepackage{algorithm}%
\usepackage{algorithmicx}%
\usepackage{algpseudocode}%
\usepackage{listings}%

\theoremstyle{thmstyleone}%
\theoremstyle{thmstyletwo}%

\theoremstyle{thmstylethree}%

\newcommand{\chgU}[1]{\textcolor{black}{#1}}

\newcommand{\chgR}[1]{\textcolor{black}{#1}}

\makeatother
\begin{document}

\title[Article Title]{Deep Learning Based Superconductivity Prediction and Experimental Tests}


\author[1]{\fnm{Daniel} \sur{Kaplan}}\email{d.kaplan1@rutgers.edu}

\author[2]{\fnm{Adam} \sur{Zheng}}\email{dzheng\_box@hotmail.com}

\author[3]{\fnm{Joanna} \sur{Blawat}}\email{jblawat@email.sc.edu}

\author[3]{\fnm{Rongying} \sur{Jin}}\email{rjin@mailbox.sc.edu}

\author[4]{\fnm{Robert J.} \sur{Cava}}\email{rcava@princeton.edu}

\author[1]{\fnm{Viktor} \sur{Oudovenko}}\email{udo@physics.rutgers.edu}

\author[1]{\fnm{Gabriel} \sur{Kotliar}}\email{kotliar@physics.rutgers.edu}

\author*[1,5,6]{\fnm{Anirvan M.} \sur{Sengupta}}\email{anirvans.physics@gmail.com}

\author*[2]{\fnm{Weiwei} \sur{Xie}}\email{xieweiwe@msu.edu}

\affil[1]{\orgdiv{Department of Physics and Astronomy}, \orgname{Rutgers University}, \orgaddress{\street{136 Frelinghuysen Road}, \city{Piscataway}, \postcode{08854}, \state{NJ}, \country{USA}}}

\affil[2]{\orgdiv{Department of Chemistry}, \orgname{Michigan State University}, \orgaddress{\street{578 S Shaw Lane}, \city{East Lansing}, \postcode{48824}, \state{Michigan}, \country{USA}}}

\affil[3]{\orgdiv{Department of Physics and Astronomy}, \orgname{University of South Carolina}, \orgaddress{\street{516 Main Street}, \city{Columbia}, \postcode{29208}, \state{South Carolina}, \country{USA}}}

\affil[4]{\orgdiv{Department of Chemistry}, \orgname{Princeton University}, \orgaddress{\street{Washington Road}, \city{Princeton}, \postcode{08544}, \state{New Jersey}, \country{USA}}}

\affil[5]{\orgdiv{Center for Computational Quantum Physics}, \orgname{Flatiron Institute}, \orgaddress{\street{162 5th Avenue}, \city{New York}, \postcode{10010}, \state{New York}, \country{USA}}}

\affil[6]{\orgdiv{Center for Computational Mathematics}, \orgname{Flatiron Institute}, \orgaddress{\street{162 5th Avenue}, \city{New York}, \postcode{10010}, \state{New York}, \country{USA}}}


\abstract{The discovery of novel superconducting materials is a longstanding
challenge in materials science, with a wealth of potential for
applications in energy, transportation, and computing. Recent
advances in artificial intelligence (AI) have enabled
expediting the search for new materials by
efficiently utilizing vast materials databases. In this study, we developed an approach based on deep learning (DL) to predict new superconducting materials. We have synthesized \chgU{a} compound derived from our DL network and confirmed its superconducting properties in agreement with our prediction.
Our approach is also compared to previous work based on random forests (RFs). In particular, RFs require knowledge of the chemical properties of the compound, while our neural net inputs  depend solely on the chemical composition. \chgR{With the help of hints from our network, we discover a new  ternary compound Mo\textsubscript{20}Re\textsubscript{6}Si\textsubscript{4}, which becomes superconducting below 5.4 K}. We further discuss the existing limitations and challenges associated with using AI to predict and, along with potential future research directions.}


\maketitle

\section{Introduction}
\label{introduction}

Superconductivity is a phenomenon characterized by zero electrical
resistivity and the Meissner effect -- the diamagnetic expulsion of of magnetic fields from the bulk of the sample, and is microscopically attributed to the formation of Cooper pairs,
making it an ideal system for the study of quantum
entanglement \cite{bardeen1957theory,gui2021chemistry,bean1964magnetization}. The search for new superconducting
materials has been a long-standing challenge in materials science with potential applications in energy, transportation, and
computing  \cite{norman2011challenge,sun2019high,foltyn2007materials}. Traditional methods of discovering new superconductors involve extensive experimental trial and error, which
can be time-consuming and expensive. In recent years, researchers have explored the use of machine learning algorithms to accelerate the discovery of new superconducting materials \cite{avery2019predicting,lookman2019active,hoffmann2022superconductivity,meredig2018can}.

Modern machine learning techniques, such as deep learning \cite{lecun2015deep}, combined with genetic algorithms \cite{david2014genetic}, can efficiently screen
large databases of materials properties and predict the properties of
potential new materials \cite{pilania2013accelerating,peterson2021materials,stanev2021artificial}. Beyond identifying potential materials, machine learning has also emerged as a tool for providing
researchers with a deeper understanding of certain
properties. For example, machine learning was
used to develop simple models that explain the electronic structures of
half-Heusler phases for potential thermoelectric
properties \cite{dylla2020machine}, or direct the search for
compounds with high hardness \cite{mazhnik2020application} by predicting the elastic moduli as a proxy \cite{mansouri2018machine}, and so on. 

In this work, we evaluated the effectiveness of deep learning methods to classify materials into superconductor/non-superconductors and also to predict $T_c$ when it is a superconductor. Our tools were applied to predict superconducting properties of hypothetical materials in sigma phases and compared to experimental findings. On this very special and small set of materials, the performance of both the random forest method and our deep learning networks are mixed. Still, we believe that we are at the beginning of the integration of AI and experimental synthesis, and this integration has immense potential for expediting the discovery of new superconductors and facilitating a deeper understanding
of the underlying superconductivity phenomena.

\section{Relationship to Other Works on Predicting Superconductivity and Superconductors}
\subsection{Evolutionary Algorithms}
While evolutionary algorithms have been in use for materials design and discovery \cite{chakraborti2004genetic,jennings2019genetic}, their application to superconductivity search is still limited.
Genetic Algorithm and Genetic Programming have been used for searching superconducting hydrogen compounds \cite{ishikawa2019materials}. We focus here on improving supervised learning approaches to predict superconductivity and remain flexible on the strategy to search for new materials.

\subsection{\chgR{Tree Ensemble Methods with Descriptors Based on Elemental Properties}}
Supervised learning methods like random forests \chgU{have} been used to predict superconductivity of materials from their chemical composition \cite{stanev2018machine} \chgU{with associated features}.  
In \citet{stanev2018machine}, one uses detailed chemical knowledge about properties of the constituent atoms, using the Materials Agnostic Platform for Informatics and Exploration (Magpie) descriptors \cite{ward2016general}. Our deep-learning-based methods \chgU{become} comparable \chgU{in} performance, needing only the  \chgU{relative atomic composition}. Thus, we expect that the internal representation of our network acquired characteristics which substitute detailed chemical knowledge. \chgR{Gradient boosting has been employed in Ref. \cite{hamidieh2018data} to predict $T_c$ of superconductors.  The input requires information on thermal conductivity, atomic radii, valence, electron affinity, and atomic
mass of the constituents, and the method cannot determine whether a material superconducts.}

\subsection{Deep Learning}
Recent work on deep learning in the context of $T_c$ estimation was carried out by Konno \textit{et al.} \cite{konno2021deep}. In this work, it was proposed that the periodic table may be learned by dividing the elements into blocks of dominant valence (i.e., s,p,d,f). By adding convolutional layers, the relationship between $T_c$ and the valence structure is learned. We note that Ref.~\cite{konno2021deep} shows better performance than the random forest model \cite{stanev2018machine} for a specifically chosen threshold $T_c$ (which determines the specificity of the model). The dataset used for learning was SuperCon~\cite{nims2011supercon}, similar to the present work. Predictions were carried on a reported materials assortment \cite{hosono2015exploration} which may include overlaps in the realm of superconductors.
Ref.~\cite{konno2021deep} stresses the importance of adding non-superconducting elements to properly eliminate overfitting and increase the rate of true positives. This is done by assigning a known database of inorganic materials (COD) \cite{COD2009} the $T_c = 0$, even though it is unknown whether all materials therein do not in fact superconduct. In the present work, we achieve similar levels of accuracy and $R^2$ in regression \textit{without} augmenting our datasets with extrinsic non-superconducting materials.
We also note that \citet{konno2021deep} dismisses usual random train-test split in favor of what they call `temporal' split. However, we note that uncertainty/error bars are not provided for Ref.~\cite{konno2021deep}'s results. 

The partitioning of the periodic table in the manner of Ref.~\cite{konno2021deep} may be thought of as adding additional features to the model. In this work, we show that a fully trained networks performs well without this additional input. Moreover, we also show that a fully connected \chgU{network} performs in a similar fashion to a convolutional network, exceeding certain expectations. \chgR{Recently, superconductor design through gradient optimization \cite{fujii2024efficientexplorationhightcsuperconductors} has been used. As to generating novel material candidates, this is a potentially useful tool. However, our data is not restricted to the high-$T_c$ regime as in Ref.~\cite{fujii2024efficientexplorationhightcsuperconductors}, but is composed of data for a wide range of superconducting temperatures.} 

\chgR{Furthermore, Deep Set architecture was used in classification in Ref. \cite{pereti2023individual}. For classification we find comparable performance relative to a more involved algorithm as presented in Ref.~\cite{pereti2023individual}. Unlike Konno \textit{et al.} \cite{konno2021deep} and Pereti \textit{et al.} \cite{pereti2023individual}, our dataset contains only real compounds extracted from SuperCon, without padding with fictitious entries for increased sensitivity. A combination of CNN and LSTM has been employed in Ref.~\cite{li2020critical}. We note that the results in the present work have lower error in prediction of $T_c$ than the best result from the hybrid neural network (HNN) in \citet{li2020critical}.} 

\section{Deep Learning Superconductivity Prediction}
\subsection{Data and Preprocessing}
The superconducting data for alloys were retrieved from database for superconducting materials, SuperCon \cite{nims2011supercon}. The data tabulates the experimental results of measurements of superconductivity
for 16,414 compounds. Each row indicates the \chgU{elements} in
the compound and their respective percentages (the input), as well as the measured
$T_c$ value (the output), where $T_c > 0$ indicates that the compound is superconducting. 
Table \ref{table:1} exemplifies of such data.

\begin{table}[h!]
\caption{The two examples of compounds, one is superconductive
with T\textsubscript{c} = 40.1 K and the other one is not
superconductive with T\textsubscript{c} = 0 K.}\label{table:1}%
\begin{tabular}{@{}lr@{}} 
\toprule
CHEMICAL COMPOUND & $T_c$ (K) \\ 
\midrule
Ca\textsubscript{0.4}Ba\textsubscript{1.25}La\textsubscript{1.25}Cu\textsubscript{3}O\textsubscript{6.98}
& 40.1 \\ 
Sm\textsubscript{1}Fe\textsubscript{0.8}Zn\textsubscript{0.2}As\textsubscript{1}F\textsubscript{0.2}O\textsubscript{0.8}
& 0.0 \\ 
\botrule
\end{tabular}
\end{table}

For each data point, a data vector of length 120 is generated with the index representing the atomic number of elements and its corresponding percentage being the entry for that index. These generate 16,414 sparse 120-dimensional vectors. This will be the input of our fully connected network. \chgU{In principle, it may be useful to incorporate some information about the relative position of elements in the periodic table. Several works \cite{konno2021deep, fujii2024efficientexplorationhightcsuperconductors} break up the periodic table into multiple rectangles and input the chemical compound as multiple images which is then processed by a convolutional network (CNN). We, however, reshape the 120 dimensional vector into a single $10\times12$ `image'. This approach is harder to justify, a priori, but, as we will see, our CNN seems to have slightly better performance in some cases, compared to the fully-connected network. We speculate that it benefits from bringing together chemically similar elements. Future work will explore optimal structuring of the data vector to better understanding the learned connectivities in this representation of the periodic table.}

\begin{figure}
    \centering
    \includegraphics[width=0.8\linewidth]{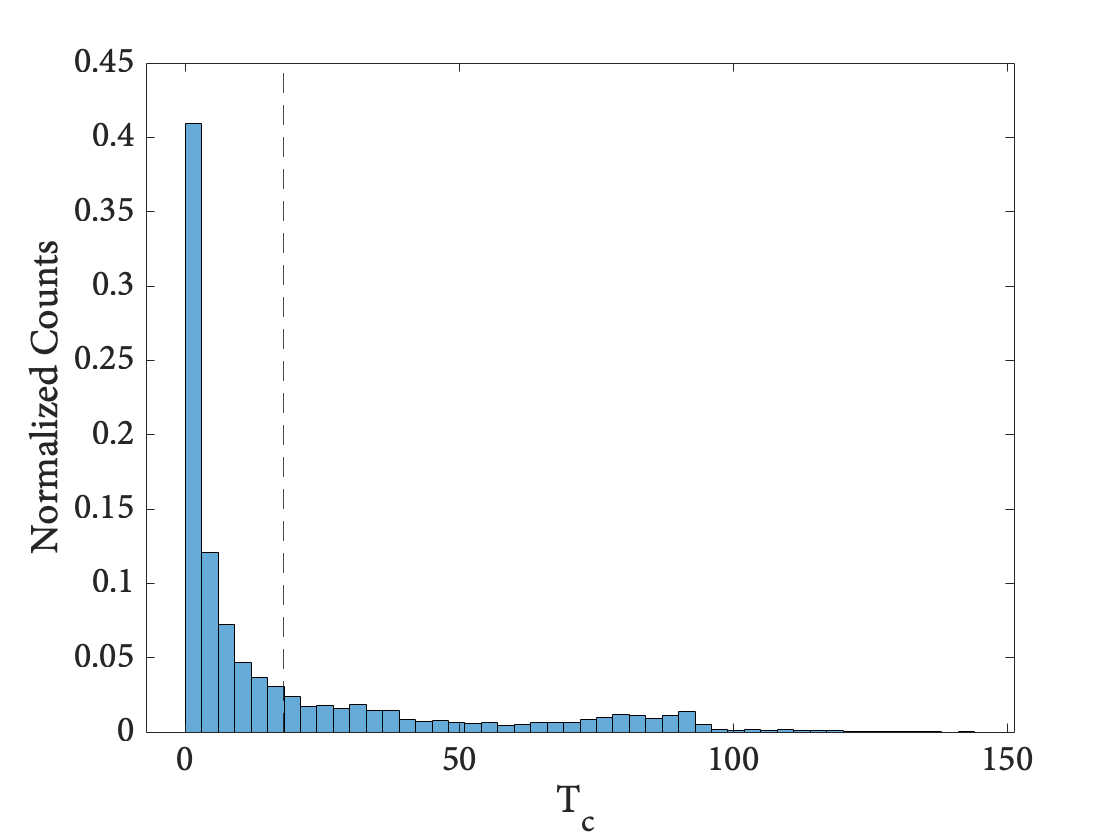}
    \caption{Histogram representing the data and $T_c$ distribution. Dashed line is the mean $T_c$ value across the dataset. $T_c$ values are in $K$.}
    \label{fig:histo}
\end{figure}

The distribution of $T_c$ in our dataset is plotted in the histogram below, Fig.~\ref{fig:histo}.

The output comprises two
components: classification ground truth, represented by a single binary
value of 0 or 1, indicating the presence or absence of
superconductivity. Thus, the two classes correspond to
non-superconducting compounds (0) and superconducting compounds (1). The
other component is the measured $T_c$ value.

\subsection{Architecture}

\begin{figure}[ht]
\centering
\includegraphics[width=0.7\columnwidth]{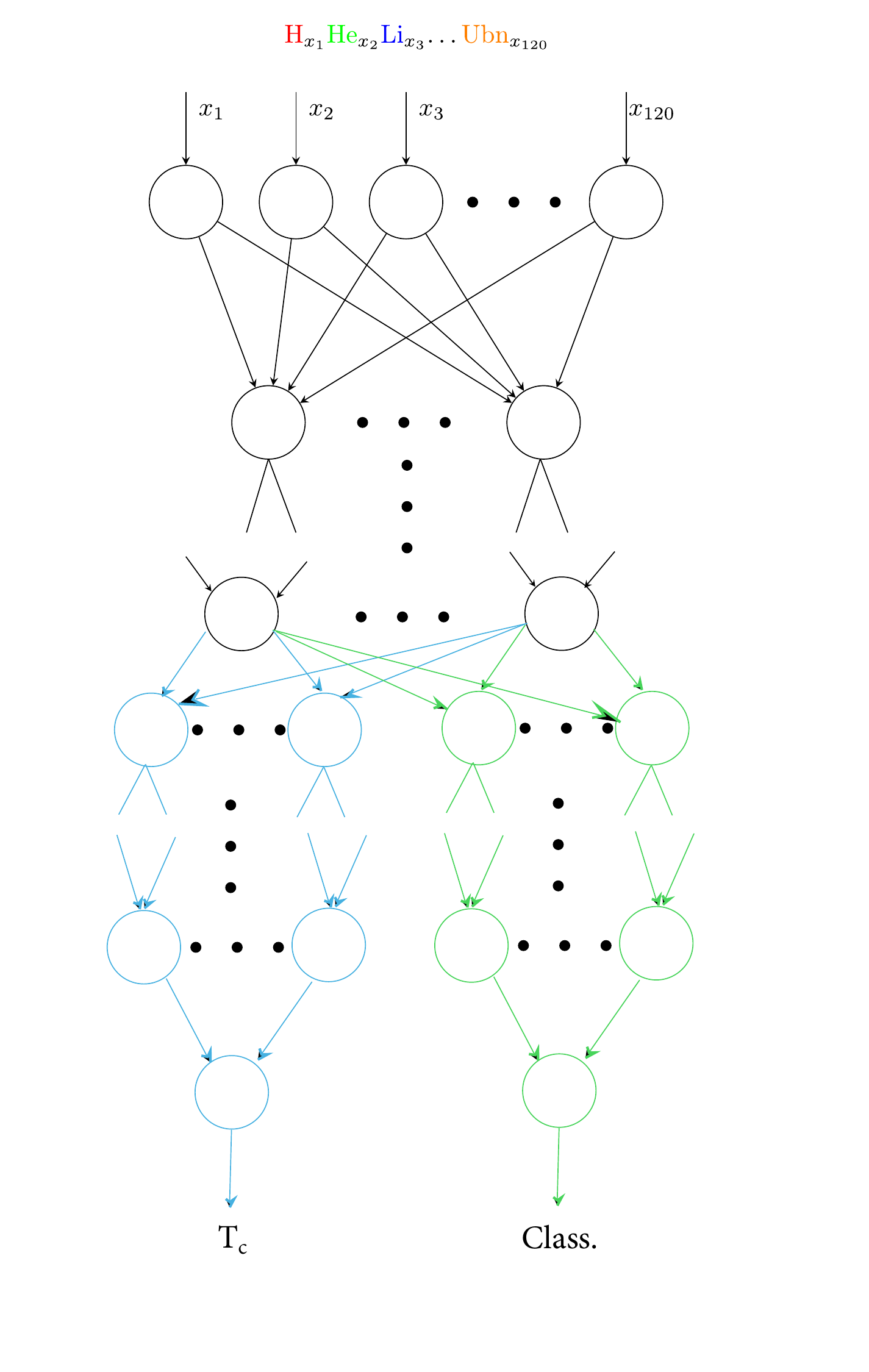}
\caption{DNN structure with a shared backbone and two prediction branches: one
for the $T_c$ value and the other for classification.}
\label{network_structure}
\end{figure}

Given the nature of the dataset, we decided to treat the problem as a
combination of classification and regression tasks. In other words, we construct deep neural nets (DNNs) that can classify
whether a given compound is superconductive or not and predict the
corresponding $T_c$ value.  The DNN model comprises three components: the
network backbone and two prediction branches: one
for classification and the other for $T_c$ value. \chgU{The architecture is schematically illustrated in Fig.~\ref{network_structure}.} We build two networks, one fully-connected, the other convolutional. For the details of the architectures, please see Supplementary Materials. The DNN model is implemented using the PyTorch \cite{paszke2019pytorch} framework,
which accepts network data in the 4-D tensor format of NC\chgU{D}HW (batch N,
channels C, depth D, height H, width W).

\chgU{We now propose justification for our choice of the branched architecture and on our planned method of training. Since the two tasks, namely, the classification of superconducting nature and the regression of $T_c$ are related, it is natural to train a set of shared latent variables as neurons in the initial layers. Given that the teaching signal from $T_c$ is more specific than the signal from the class, we train the shared part via the regression task. As the classification task is conceptually easier than the regression task, we choose to allow some additional flexibility to the network, rather than restricting it to classifiers obtained by thresholding the output of a regression network, as in Refs.~\cite{stanev2018machine,konno2021deep} and others. This allows for the obtainment of true classification, without a need to specify external thresholding criteria which skew the accuracy and recall metrics of the network.}

\subsection{Training}
\begin{figure}[ht]
\centering
\includegraphics[width=0.9\columnwidth]{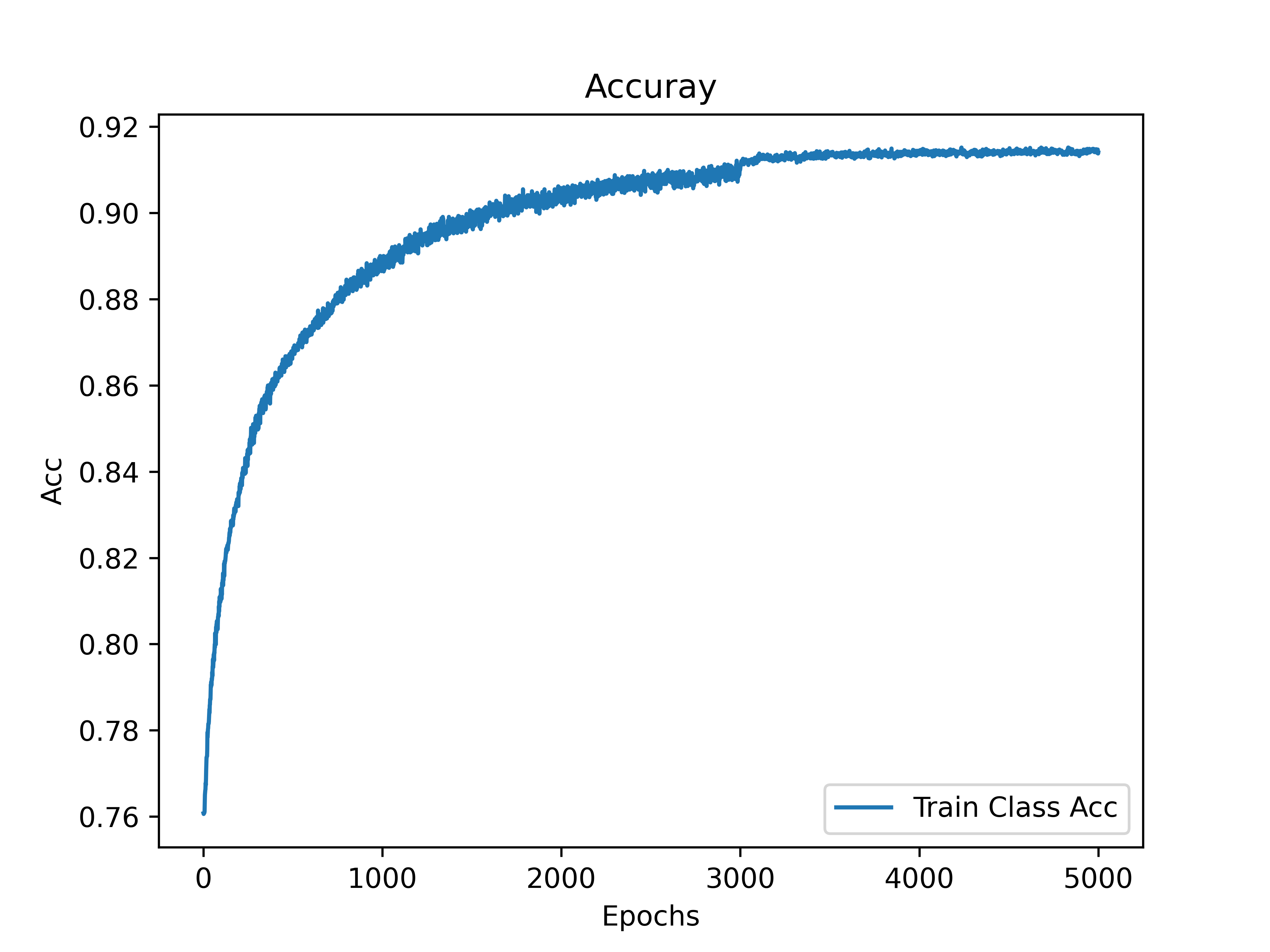}
\caption{Fully-connected Model: Accuracy in
classification during training with epoch.}
\label{fc_loss}
\end{figure}

\begin{figure}[ht]
\centering
\includegraphics[width=0.9\columnwidth]{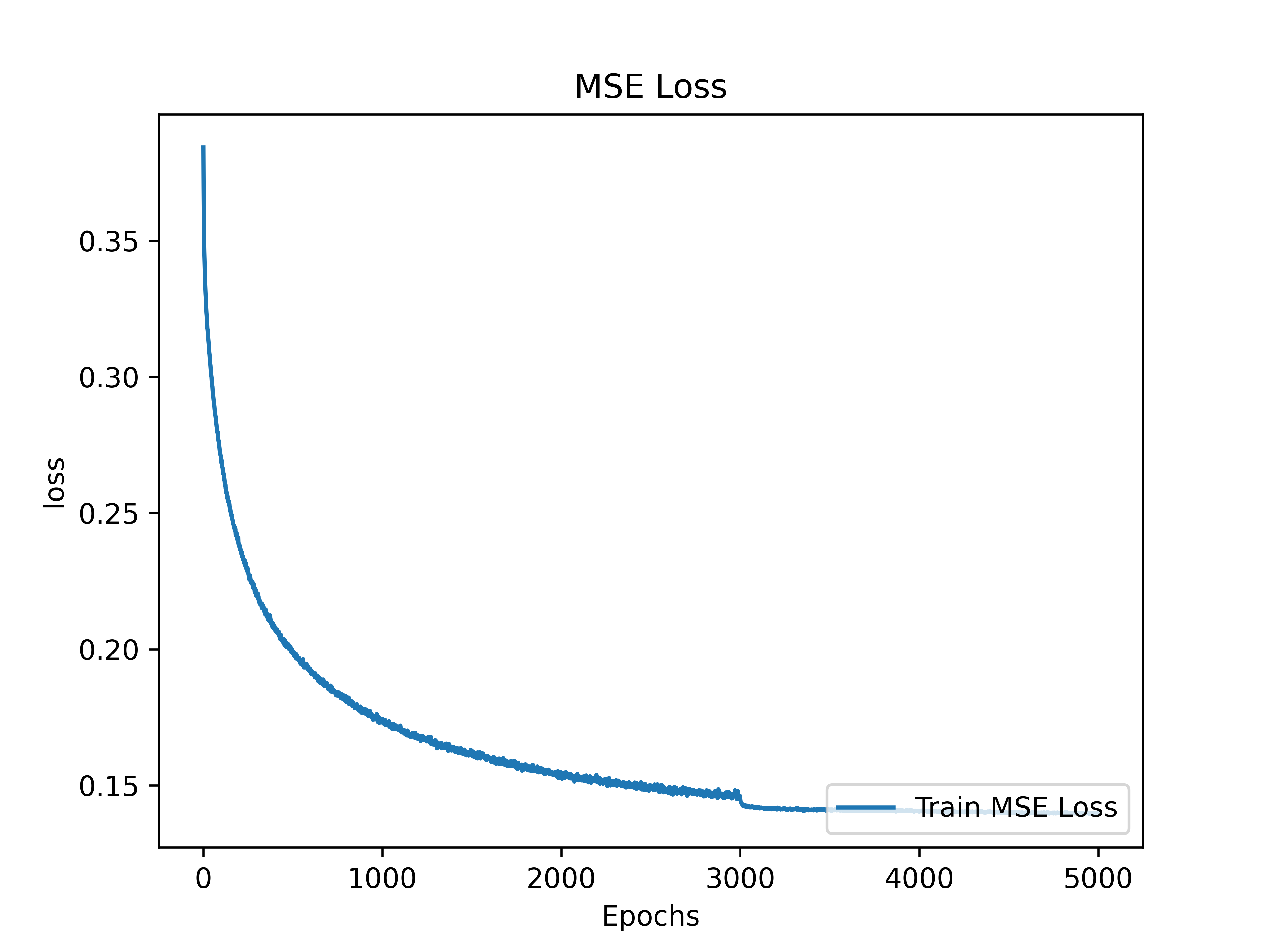}
\caption{Fully-connected Model: MSE loss in classification during training.}
\label{fc_acc}
\end{figure}

\begin{figure}[ht]
\centering
\includegraphics[width=0.9\columnwidth]{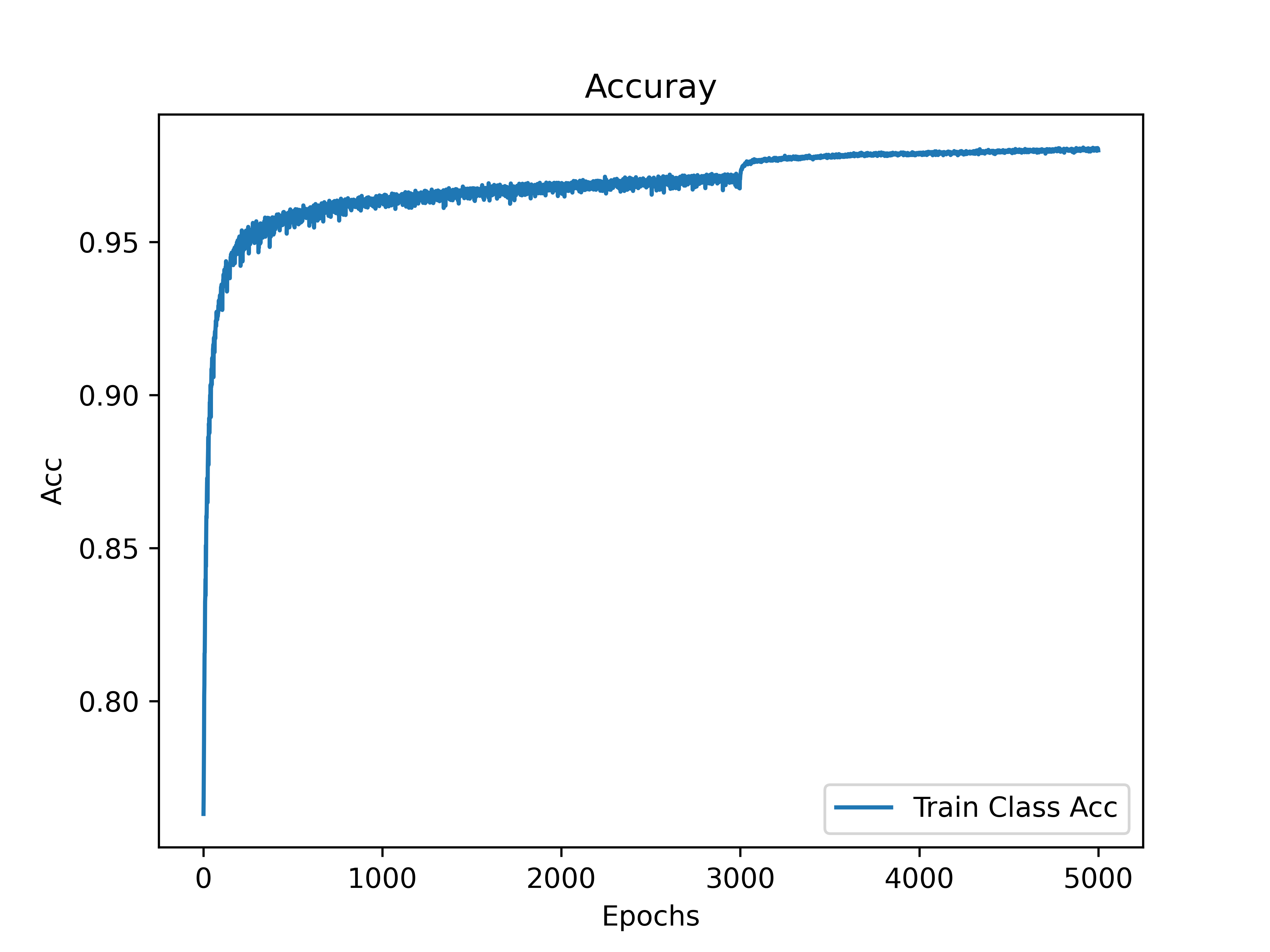}
\caption{Convolutional Model: Accuracy in
classification during training with epoch.}
\label{cnn_acc}
\end{figure}

\begin{figure}[ht]
\centering
\includegraphics[width=0.9\columnwidth]{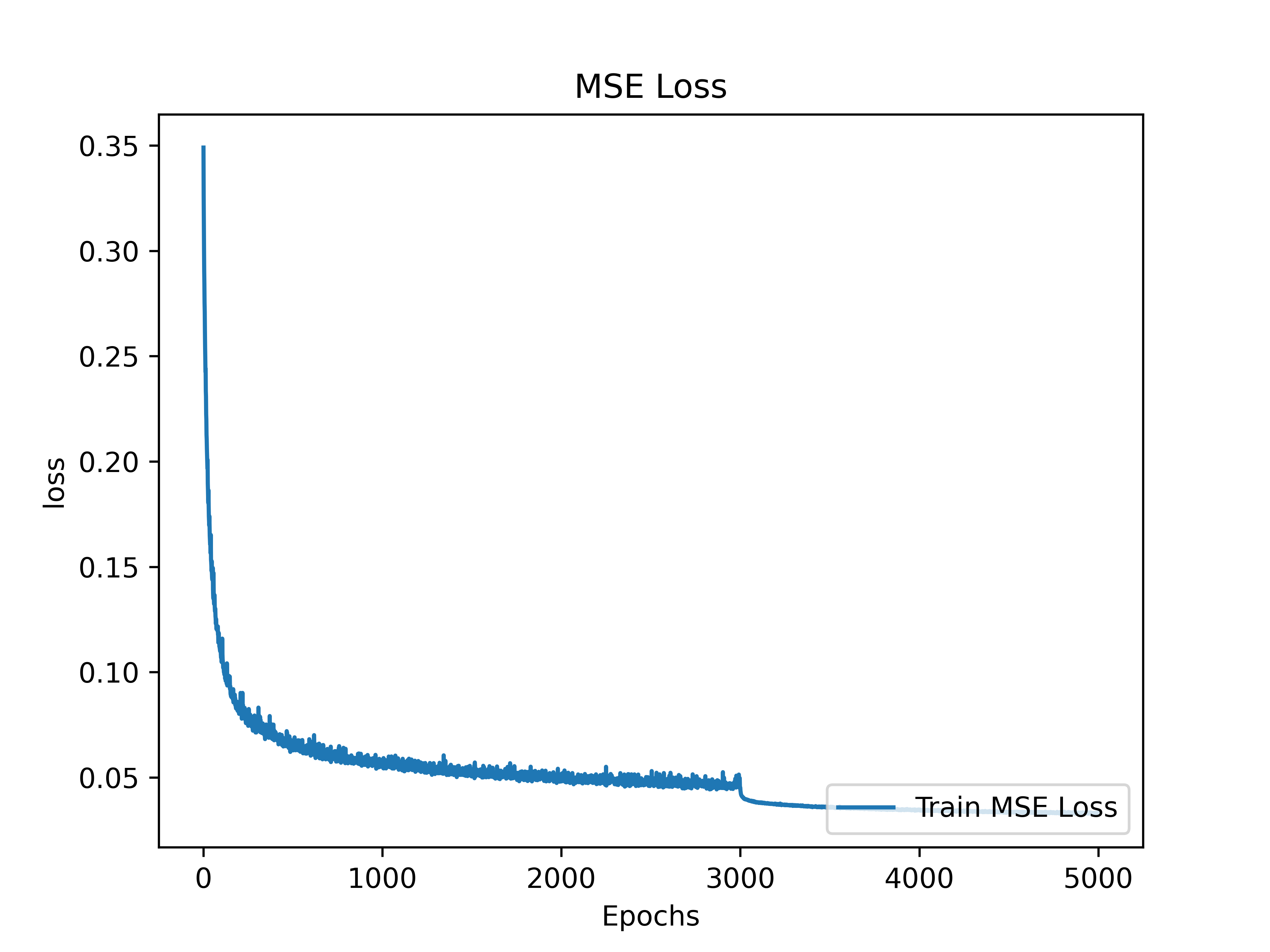}
\caption{Convolutional Model: MSE loss in
classification during training with epoch.}
\label{cnn_loss}
\end{figure}

The data set is divided into 2 portions:
training (13132 entries) and testing dataset (3282 entries). \chgR{One common practice to prevent overfitting is to carry out multiple random test-train splits, which provides us with error bars on our estimates of the test performance}. \chgR{In our study, we have used six test-train splits, for each model}.  
During the training process, the Mean Squared Error (MSE) is utilized to compute the LOSS error, which is subsequently leveraged for the backpropagation
calculation. The training process is carried out in
two stages. First, the Backbone and $T_c$ Prediction
branches are jointly trained, and the loss of the predicted
$T_c$ value versus the ground truth is used for network
backpropagation. Once the T\textsubscript{c} Prediction results reach a satisfactory level, the parameters of the Backbone and
T\textsubscript{c} Prediction branches are fixed. The training is then performed solely on the Classification branch. Each branch was run for 5,000 epochs. This procedure was performed for both architectures. \chgU{The learning curves, showcasing performance in accuracy and MSE loss are plotted in Figs.~\ref{fc_acc}, \ref{fc_loss}, \ref{cnn_acc}, and, \ref{cnn_loss}.}


After completing Stage1 and Stage2 training, the model is finalized for testing. The training process is further optimized by using two-step learning rate decay. The initial learning rate is set to 0.0001, and at 3000 epochs, it is decayed to 0.00001 to fine-tune the training parameters. The training results are evaluated based on the backbone and
T\textsubscript{c} prediction branch training, and the
T\textsubscript{c} prediction difference is calculated by averaging the
mean absolute error (MAE) difference between the predicted and
ground-truth values \cite{chai2014root}. A lower difference value
indicates more accurate prediction results. Both the loss and
T\textsubscript{c} prediction differences decrease with an increase in
epoch count, and starting from the 2000\textsuperscript{th} epoch, the
loss and difference curve reach a plateau. However, at the
3000\textsuperscript{th} epoch, with a decay in the learning rate from
0.0001 to 0.00001, the loss and T\textsubscript{c} prediction
differences further decrease, indicating that the learning rate decay
indeed helps to improve the training process by providing finer granularity control of parameter updating.

\subsection{Results}

Results and the
performance of the classification branch is evaluated based on its
accuracy in correctly identifying if a compound exhibits
superconductivity or not, as compared to the ground-truth. A higher
accuracy value indicates better performance in the classification task.
The testing process involves evaluating the performance of the trained
model on a separate dataset known as the test dataset, which is distinct
from the data used during model training. The test dataset comprises a
total of 3282 entries. The final training and testing results are listed
in Tables~\ref{table:2}-\ref{table:2_1}.

\begin{table}[h!]
\caption{The summary of the training and testing results for the fully connected network with average prediction difference, average $T_c$ and classification accuracy.}\label{table:2_1}%
\begin{tabular}{@{}lll@{}} 
 \toprule
  & Avg pred. diff. (K) & Classification \\
    &  / avg $T_c$ value (K) &  \\ 
\midrule
\textbf{Training} & 2.600 $\pm$ 0.194 / 17.9792 & 91.95 $\pm$ 0.3\%\\ 
\textbf{Testing} & 4.497 $\pm$ 0.328 / 17.9195 & 83.04 $\pm$ 0.6 \% \\ 
\botrule
\end{tabular}
\end{table}

\begin{table}[h!]
\caption{The summary of the training and testing results for the convolutional network with average prediction difference, average $T_c$ and classification accuracy.}\label{table:2}%
\begin{tabular}{@{}lll@{}} 
 \toprule
  & Avg pred. diff. (K) & Classification \\
    &  / avg $T_c$ value (K) &  \\ 
\midrule
\textbf{Training} & 1.470 $\pm$ 0.047 / 17.979 & 97.97 $\pm$ 0.1\%\\ 
\textbf{Testing} & 3.208 $\pm$ 0.180 / 17.9195 & 84.77 $\pm$ 0.3\% \\ 
\botrule
\end{tabular}
\end{table}

Comparison of our two models with \citet{stanev2018machine} for classification performance is in Table \ref{table:4}.


Tab. ~\ref{table:4} summarizes our test results comparing the baseline RF (Ref. \cite{stanev2018machine}) with our models
\begin{table}[h!]
\caption{The summary of classification and regression test results for various methods. RF= Random Forest \cite{stanev2018machine}, FCNN= Fully-connected neural net, CNN= Convolutional Neural Net. The error in $T_c$ regression (Reg.) is defined as the mean of $T_c(\textrm{pred.})-T_c(\textrm{actual})$. }\label{table:4}%
\begin{tabular}{@{}lllllll@{}} 
 \toprule
 Model & Accuracy & Precision & Recall & F1 & Reg. [K] & Class. \\
 \midrule
 \textbf{RF} \cite{stanev2018machine} & 85\% \footnotemark  & 85\% & 94\% & 90\% & 7.88 $\pm$ 0.52  & NA \footnotemark \\ 
\hline
\textbf{FCNN} & 83 $\pm$ 0.6 \%  & 85 $\pm$ 0.7\% & \textbf{95 $\pm$ 0.4\%} & 90\% & 4.497 $\pm$ 0.328 & 83.04 $\pm$ 0.6\%  \\ 
 \hline
\textbf{CNN} & 85 $\pm$ 0.3\% & \textbf{88 $\pm$ 0.4\%} & 92 $\pm$ 0.3\% & 90\% & \textbf{3.208 $\pm$ 0.180} & \textbf{84.77 $\pm$ 0.3\%} \\  
\bottomrule
\end{tabular}
\end{table}

\let\thefootnote\relax\footnotetext{\textsuperscript{1} To match with our definition of accuracy, this value was extrapolated from Stanev \textit{et al.} to the point $T_{sep} = 0$.}
\let\thefootnote\relax\footnotetext{\textsuperscript{2} Stanev \textit{et al.} do not explicity train for classification, we do not present a direct comparison as it would be fine-tuned by their value of $T_{sep}$.}

The SuperCon database \cite{nims2011supercon} has 16,414 entries, 12,499 ($\approx76\%$) of which are known to be superconductors and the rest are considered non-superconducting. Note that a positive/negative weight guessing random classifier  could achieve a maximum of $\approx 76\%$ accuracy by always going with the majority class \cite{brownlee2019probability}. In this dataset, the majority class predictor would have $100\%$ recall  but $\approx 76\%$ precision. These baselines are to be compared with the classification performances of the machine learning methods.

\section{Experimental Studies}
Superconductivity has been observed in a few materials belonging to the $\sigma$-phase category \cite{carnicom2017superconductivity}, characterized by a
high degree of disorder and the presence of multiple elements occupying
the same Wyckoff position. The occurrence of $\sigma$-phases is strongly
dependent on the concentration and the type of the constituent elements. Notably,
superconductivity has been demonstrated in materials such as Nb-Ru-Ge,
Nb-Rh-Ge, and Nb-Rh-Si, which has stimulated further investigations into
Re-based materials with varying ternary systems to explore the
superconducting properties of disordered Re-based
materials \cite{carnicom2017new,carnicom2018sigma}. 

We leveraged our chemical intuition and expertise, to narrow our search for new superconductors to ternary compounds with formula unit Mo\textsubscript{20}X\textsubscript{6}Z\textsubscript{4}, with X=Re, Rh, Ru, Z=Ge, Si, as the 20-6-4 stoichiometry is most likely to form in the sigma phase. Then we use our
DNN-based algorithm to investigate the potential
superconducting properties of these compounds.

Our CNN predicted that
``Mo\textsubscript{4}Re\textsubscript{2}Si" is likely to exhibit
superconductivity with a critical temperature ($T_c$) of
approximately 6 K, while ``Mo\textsubscript{4}Re\textsubscript{2}Ge" is
also predicted to be a superconductor with $T_c\sim5$ K. 

We now discuss possible data leakage \cite{cook2023data} concerning these predictions. Data leakage occurs when the training data contains some kind of information that will not be available when the model is used for prediction or decision making. Such leakage has two main types: target leakage and train-test contamination. Target leakage occurs when training predictors include data that will not be available at the time of making new predictions. That is not relevant in the context of this work. As to test-train contamination, there was nothing in our dataset of the form Mo\textsubscript{x}Re\textsubscript{y}Si\textsubscript{z}. 

If one broadens the definition and considers similar materials, our dataset indeed has materials which share two of the three elements present in the target. While these materials, present in the dataset could be similar, for example, some materials of the form Mo\textsubscript{x}Re\textsubscript{y}O\textsubscript{z}, superconductivity is dependent on many subtle effects and is not easy to predict. In fact, different compositions of Mo\textsubscript{x}Re\textsubscript{y}O\textsubscript{z}, have different behavior. In the course of predicting new superconducting materials, it is natural to expect superconducting would occur in compounds with relatively similar composition. Thus, for computationally accessible new superconducting materials, some similarity in the dataset is expected and warranted.

Our experimental results show that the compound obtained by starting from the loading composition Mo\textsubscript{4}Re\textsubscript{2}Si is a superconductor with
T\textsubscript{c} at 5.4 K. Thus, the machine learning approach assisted chemical
intuition in finding a new superconductor. However, no superconductivity was observed above 1.8 K in the material obtained by starting from the loading composition 
Mo\textsubscript{4}Re\textsubscript{2}Ge. For clarification, by starting with the loading composition of 4-2-1 ratio, we get final products with around 20-6-4 ratio.

\subsection{Superconductivity in
Mo\textsubscript{20}Re\textsubscript{6}Si\textsubscript{4}}
\label{sec:experiment}
\begin{figure}[ht]
\centering
\includegraphics[width=0.9\columnwidth]{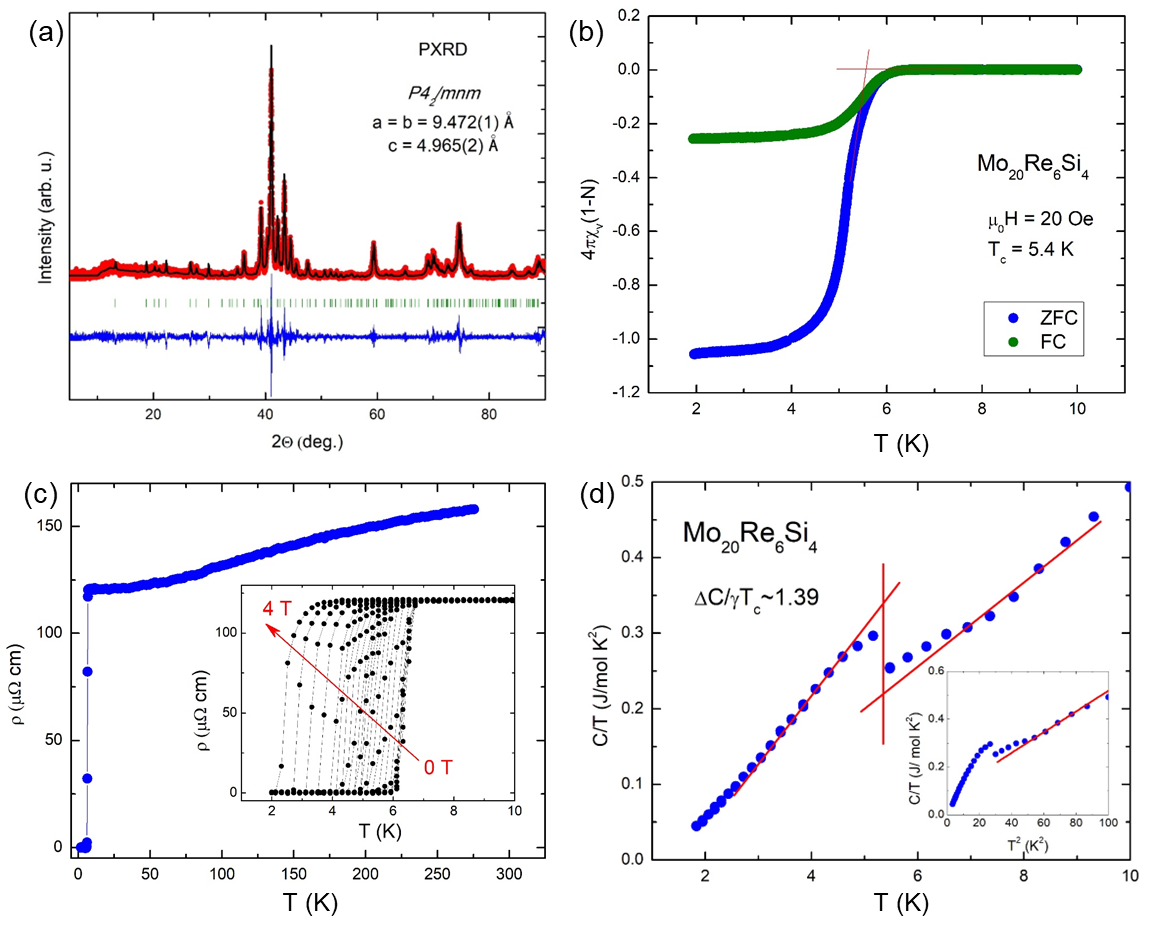}
\caption{Phase identification and superconductivity properties
of Mo\textsubscript{20}Re\textsubscript{6}Si\textsubscript{4} (a)
Powder X-ray diffraction pattern of sigma phase
Mo\textsubscript{20}Re\textsubscript{6}Si\textsubscript{4}; \chgR{The diffraction pattern refinement confirms that this is a member of $\sigma$-phase family, possessing a tetragonal structure.} (b) Magnetic
susceptibility of
Mo\textsubscript{20}Re\textsubscript{6}Si\textsubscript{4}. \chgR{The sharp drop at zero-field cooling (ZFC) $ T = 5.4 \textrm{K}$ is indicative of bulk superconductivity. It is to be contrasted with the field-cooled (FC) result.}  (c) \chgR{Resistivity as a function of temperature, confirming superconducting at $T = 5.4 \textrm{K}$, where the resistance drops to zero. Inset shows the}
Field-dependent resistivity of
the compound, \chgR{showing bulk superconductivity}. (d) Heat
capacity of Mo\textsubscript{20}Re\textsubscript{6}Si\textsubscript{4}. \chgR{The discontinuity in specific heat demonstrating bulk superconductivity. Inset shows the two regimes, above and below $T_c$: above $T_c$ phonon dominated normal heat capacity. Below $T_c$, a strong suppression of $C_p/T$, consistent with the presence of a bulk gap. }
\chgR{For further details see main text, Subsec.~\ref{sec:experiment}.}
}
\label{experimental}
\end{figure}

Mo\textsubscript{20}Re\textsubscript{6}Si\textsubscript{4} compound was
synthesized successfully by arc melting method, and its crystal
structure was confirmed to be tetragonal with lattice parameters
\emph{a} = \emph{b} = 9.472(1) Å and \emph{c} = 4.965(2) Å. The compound
belongs to the $\sigma$-phase, which has a large unit cell with 30 atoms and a
general stoichiometry of A\textsubscript{20}B\textsubscript{10} shown in
Figure \ref{experimental}a. In binary $\sigma$-phases, A and B atoms occupy
different Wyckoff positions, and in ternary $\sigma$-phases, the third element
partially occupies some positions from A and/or B atoms. Although
further experiments are required to determine all mixing and site
occupancies, assuming the stoichiometry of
Mo\textsubscript{20}Re\textsubscript{6}Si\textsubscript{4} can be used
in further physical properties analysis. 

Figure \ref{experimental}b displays the temperature dependence of the zero-field-cooled (ZFC) and field-cooled (FC) volume magnetic susceptibility measured under an
applied magnetic field of 20 Oe. The transition to the superconducting
state is observed at T\textsubscript{c} = 5.4 K and was determined as
the point of intersection between the steepest slope of
$\chi_{v}$ and the extrapolation of the normal state
susceptibility. This method is consistent with previous studies. The ZFC
volume magnetic susceptibility slightly exceeds $4\pi \chi_v$ = -1, which is the
expected value for a perfect Meissner effect. 

In order to gain a deeper understanding of the superconducting state of the
material, we conducted resistivity measurements. The temperature
dependence of the resistivity is depicted in Figure \ref{experimental}c. We
observed a decrease in resistivity as the temperature increased, which
is typical for metallic behavior. However, the resistivity values were
found to be high, likely due to the high level of disorder present in
the material. Notably, a sharp drop to zero resistivity was observed at
a temperature of approximately 6 K, which is indicative of the
superconducting transition. 

Finally, to establish the bulk nature of the observed superconductivity, we
performed specific heat measurements. The temperature dependence of C/T
is presented in Figure \ref{experimental}d. We observed a $\lambda$-shaped anomaly,
which indicates the presence of a phase transition. To identify the
specific heat jump ($\Delta$C) and critical temperature Tc, we constructed an
equal-area entropy plot (red lines in the main panel of Figure \ref{experimental}d). The obtained Tc was determined to be 5.4 K, which is in
agreement with the magnetic susceptibility data.

\section{Conclusion}

\label{conclusion}
This work proposes a DNN-based superconductivity prediction model that uses 13132 data points to achieve an accuracy of 84\% on 3282 test data. This is a small dataset compared to many cutting edge applications of DNNs. For instance, popular image classification algorithms such as mobile-net and res-net that use
ImageNet as a basis for training are trained on 1.2 million images to
classify 1000 objects. The top-1 accuracy for ImageNet has moved into 90\% plus territory only within the last few years (see \href{https://paperswithcode.com/sota/image-classification-on-imagenet}{report}).

Our performance is comparable to previous work  using random forests, but does not require detailed atomic-level chemical information utilized by \citet{stanev2018machine} while using random forests.
\chgR{Unlike Konno \textit{et al.} \cite{konno2021deep} and Pereti \textit{et al.} \cite{pereti2023individual}, our dataset contains only real compounds extracted from SuperCon, without padding with fictitious entries for increased sensitivity. Our data is further not restricted to the high-$T_c$ regime as in Ref.~\cite{fujii2024efficientexplorationhightcsuperconductors}, but is composed of data for a wide range of superconducting temperatures. In classification we find comparable performance relative to a more involved algorithm as in Ref.~\cite{pereti2023individual}.}

To enhance the prediction accuracy and robustness of the model, more training data is required. Also, the our training data set only encompasses the chemical composition of the material. \chgU{One extension is to encode the neighborhood of elements in the periodic table, as in the graph sometimes known as the ``periodic spiral" \cite{rodriguez2024periodic}. } Meanwhile, the sigma phase experiments
indicate that all the methods have difficulties differentiating between
elements in the same group. Utilizing the crystal structure \cite{sommer20233dsc} and
electronic structure can provide additional information to the model to
abstract during the training process. Also interesting is the use of symmetries for prediction \cite{tang2022high}. These augmentations of predictors for the supervised algorithms hopefully boosts the
model's robustness, improves its ability to generalize
information, and make predictions in real-world scenarios.

\chgU{\section{Acknowledgements} 
DK acknowledges the support of the Abrahams Postdoctoral Fellowship of the Center for Materials Theory, Rutgers University, and the Zuckerman STEM Fellowship.
GK acknowledges the support of NSF DMR-1733071. WX's work is supported by U.S.DOE-BES under Contract DE-SC0023648.}

\section{Data Availability} The original SuperCon data used in this study are available at the references given in this paper \cite{nims2011supercon}. Datasets generated during the current study are available from the corresponding author on reasonable request. 
\chgU{The code used in this work is freely available at: \href{https://github.com/danielkaplan137/TcMLPred}{TcMLPred}. }

\begin{appendices}

\section{Details of Architecture etc.}\label{secA1}

The convolutional model uses both convolution and pooling with the later stages being fully connected. The fully-connected model obviously uses all to all connections at every layer. The activation function used for both network is ReLU. We provide the visualization of the networks using the Netron software (\url{https://netron.app/}), based on the saved .onnx files.

\begin{figure}[ht]
\centering
\includegraphics[width=.5\columnwidth]{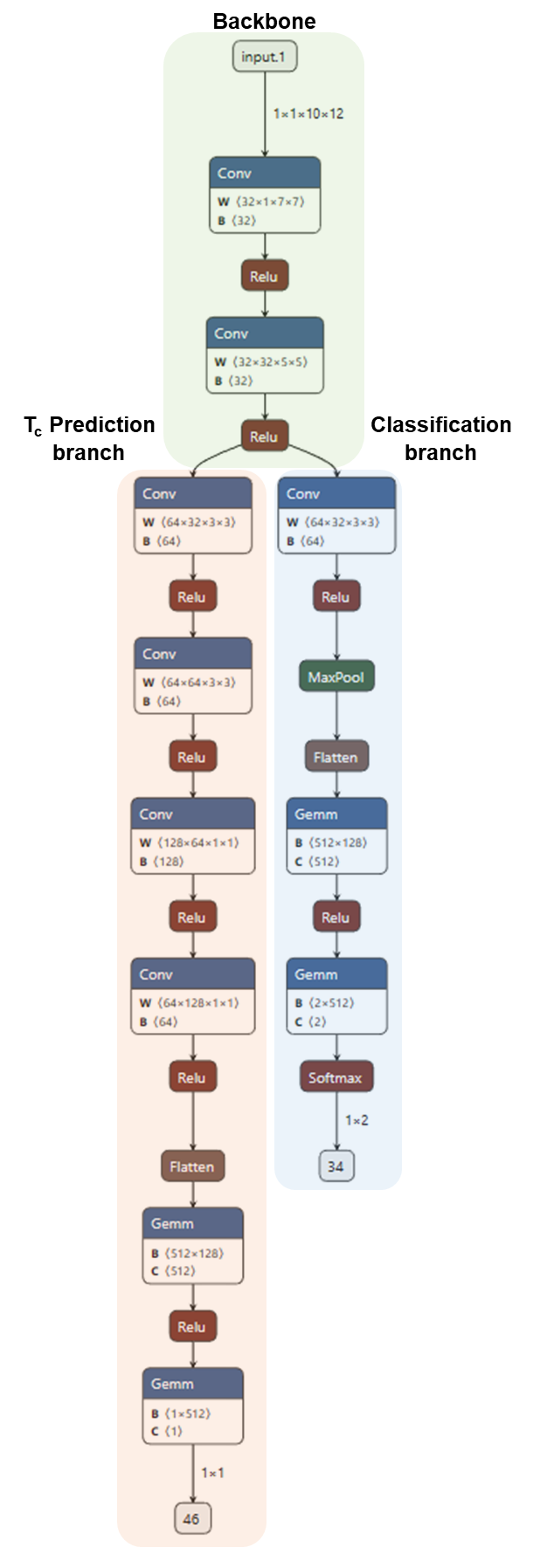}
\caption{The convolutional model, including (1)
Backbone, (2) T\textsubscript{c} Prediction branch and (3)
Classification branch.}
\label{convolutional}
\end{figure}

\begin{figure}[ht]
\centering
\includegraphics[width=0.40\columnwidth]{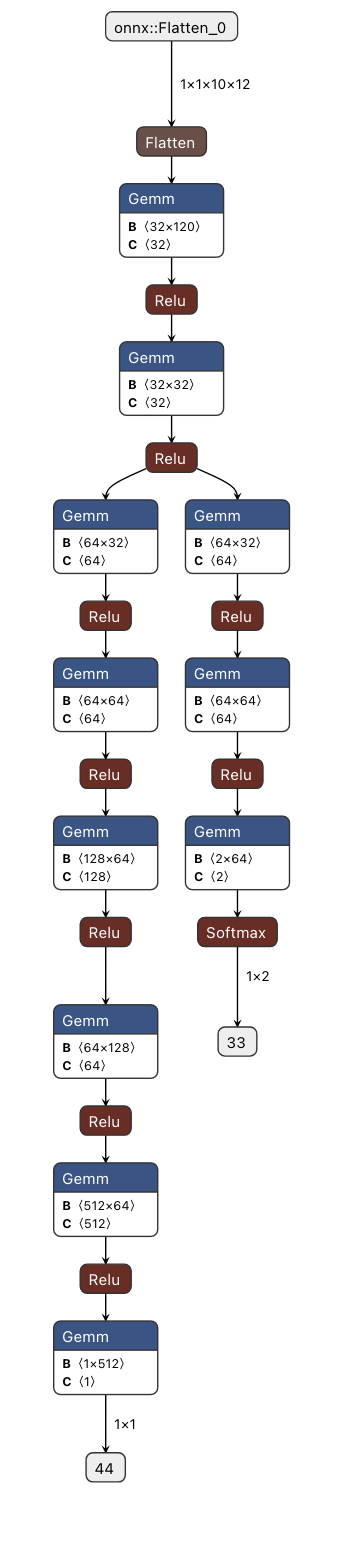}
\caption{The fully-connected model.}
\label{fully_conn}
\end{figure}

\end{appendices}

\bibliography{main.bbl}


\begin{thebibliography}{41}
\ifx \bisbn   \undefined \def \bisbn  #1{ISBN #1}\fi
\ifx \binits  \undefined \def \binits#1{#1}\fi
\ifx \bauthor  \undefined \def \bauthor#1{#1}\fi
\ifx \batitle  \undefined \def \batitle#1{#1}\fi
\ifx \bjtitle  \undefined \def \bjtitle#1{#1}\fi
\ifx \bvolume  \undefined \def \bvolume#1{\textbf{#1}}\fi
\ifx \byear  \undefined \def \byear#1{#1}\fi
\ifx \bissue  \undefined \def \bissue#1{#1}\fi
\ifx \bfpage  \undefined \def \bfpage#1{#1}\fi
\ifx \blpage  \undefined \def \blpage #1{#1}\fi
\ifx \burl  \undefined \def \burl#1{\textsf{#1}}\fi
\ifx \doiurl  \undefined \def \doiurl#1{\url{https://doi.org/#1}}\fi
\ifx \betal  \undefined \def \betal{\textit{et al.}}\fi
\ifx \binstitute  \undefined \def \binstitute#1{#1}\fi
\ifx \binstitutionaled  \undefined \def \binstitutionaled#1{#1}\fi
\ifx \bctitle  \undefined \def \bctitle#1{#1}\fi
\ifx \beditor  \undefined \def \beditor#1{#1}\fi
\ifx \bpublisher  \undefined \def \bpublisher#1{#1}\fi
\ifx \bbtitle  \undefined \def \bbtitle#1{#1}\fi
\ifx \bedition  \undefined \def \bedition#1{#1}\fi
\ifx \bseriesno  \undefined \def \bseriesno#1{#1}\fi
\ifx \blocation  \undefined \def \blocation#1{#1}\fi
\ifx \bsertitle  \undefined \def \bsertitle#1{#1}\fi
\ifx \bsnm \undefined \def \bsnm#1{#1}\fi
\ifx \bsuffix \undefined \def \bsuffix#1{#1}\fi
\ifx \bparticle \undefined \def \bparticle#1{#1}\fi
\ifx \barticle \undefined \def \barticle#1{#1}\fi
\bibcommenthead
\ifx \bconfdate \undefined \def \bconfdate #1{#1}\fi
\ifx \botherref \undefined \def \botherref #1{#1}\fi
\ifx \url \undefined \def \url#1{\textsf{#1}}\fi
\ifx \bchapter \undefined \def \bchapter#1{#1}\fi
\ifx \bbook \undefined \def \bbook#1{#1}\fi
\ifx \bcomment \undefined \def \bcomment#1{#1}\fi
\ifx \oauthor \undefined \def \oauthor#1{#1}\fi
\ifx \citeauthoryear \undefined \def \citeauthoryear#1{#1}\fi
\ifx \endbibitem  \undefined \def \endbibitem {}\fi
\ifx \bconflocation  \undefined \def \bconflocation#1{#1}\fi
\ifx \arxivurl  \undefined \def \arxivurl#1{\textsf{#1}}\fi
\csname PreBibitemsHook\endcsname

\bibitem[\protect\citeauthoryear{Bardeen et~al.}{1957}]{bardeen1957theory}
\begin{barticle}
\bauthor{\bsnm{Bardeen}, \binits{J.}},
\bauthor{\bsnm{Cooper}, \binits{L.N.}},
\bauthor{\bsnm{Schrieffer}, \binits{J.R.}}:
\batitle{Theory of superconductivity}.
\bjtitle{Physical review}
\bvolume{108}(\bissue{5}),
\bfpage{1175}
(\byear{1957})
\end{barticle}
\endbibitem

\bibitem[\protect\citeauthoryear{Gui et~al.}{2021}]{gui2021chemistry}
\begin{barticle}
\bauthor{\bsnm{Gui}, \binits{X.}},
\bauthor{\bsnm{Lv}, \binits{B.}},
\bauthor{\bsnm{Xie}, \binits{W.}}:
\batitle{Chemistry in superconductors}.
\bjtitle{Chemical Reviews}
\bvolume{121}(\bissue{5}),
\bfpage{2966}--\blpage{2991}
(\byear{2021})
\end{barticle}
\endbibitem

\bibitem[\protect\citeauthoryear{Bean}{1964}]{bean1964magnetization}
\begin{barticle}
\bauthor{\bsnm{Bean}, \binits{C.P.}}:
\batitle{Magnetization of high-field superconductors}.
\bjtitle{Reviews of modern physics}
\bvolume{36}(\bissue{1}),
\bfpage{31}
(\byear{1964})
\end{barticle}
\endbibitem

\bibitem[\protect\citeauthoryear{Norman}{2011}]{norman2011challenge}
\begin{barticle}
\bauthor{\bsnm{Norman}, \binits{M.R.}}:
\batitle{The challenge of unconventional superconductivity}.
\bjtitle{Science}
\bvolume{332}(\bissue{6026}),
\bfpage{196}--\blpage{200}
(\byear{2011})
\end{barticle}
\endbibitem

\bibitem[\protect\citeauthoryear{Sun and Cava}{2019}]{sun2019high}
\begin{barticle}
\bauthor{\bsnm{Sun}, \binits{L.}},
\bauthor{\bsnm{Cava}, \binits{R.J.}}:
\batitle{High-entropy alloy superconductors: Status, opportunities, and
  challenges}.
\bjtitle{Physical Review Materials}
\bvolume{3}(\bissue{9}),
\bfpage{090301}
(\byear{2019})
\end{barticle}
\endbibitem

\bibitem[\protect\citeauthoryear{Foltyn et~al.}{2007}]{foltyn2007materials}
\begin{barticle}
\bauthor{\bsnm{Foltyn}, \binits{S.}},
\bauthor{\bsnm{Civale}, \binits{L.}},
\bauthor{\bsnm{MacManus-Driscoll}, \binits{J.}},
\bauthor{\bsnm{Jia}, \binits{Q.}},
\bauthor{\bsnm{Maiorov}, \binits{B.}},
\bauthor{\bsnm{Wang}, \binits{H.}},
\bauthor{\bsnm{Maley}, \binits{M.}}:
\batitle{Materials science challenges for high-temperature superconducting
  wire}.
\bjtitle{Nature materials}
\bvolume{6}(\bissue{9}),
\bfpage{631}--\blpage{642}
(\byear{2007})
\end{barticle}
\endbibitem

\bibitem[\protect\citeauthoryear{Avery et~al.}{2019}]{avery2019predicting}
\begin{barticle}
\bauthor{\bsnm{Avery}, \binits{P.}},
\bauthor{\bsnm{Wang}, \binits{X.}},
\bauthor{\bsnm{Oses}, \binits{C.}},
\bauthor{\bsnm{Gossett}, \binits{E.}},
\bauthor{\bsnm{Proserpio}, \binits{D.M.}},
\bauthor{\bsnm{Toher}, \binits{C.}},
\bauthor{\bsnm{Curtarolo}, \binits{S.}},
\bauthor{\bsnm{Zurek}, \binits{E.}}:
\batitle{Predicting superhard materials via a machine learning informed
  evolutionary structure search}.
\bjtitle{npj Computational Materials}
\bvolume{5}(\bissue{1}),
\bfpage{1}--\blpage{11}
(\byear{2019})
\end{barticle}
\endbibitem

\bibitem[\protect\citeauthoryear{Lookman et~al.}{2019}]{lookman2019active}
\begin{barticle}
\bauthor{\bsnm{Lookman}, \binits{T.}},
\bauthor{\bsnm{Balachandran}, \binits{P.V.}},
\bauthor{\bsnm{Xue}, \binits{D.}},
\bauthor{\bsnm{Yuan}, \binits{R.}}:
\batitle{Active learning in materials science with emphasis on adaptive
  sampling using uncertainties for targeted design}.
\bjtitle{npj Computational Materials}
\bvolume{5}(\bissue{1}),
\bfpage{21}
(\byear{2019})
\end{barticle}
\endbibitem

\bibitem[\protect\citeauthoryear{Hoffmann
  et~al.}{2022}]{hoffmann2022superconductivity}
\begin{barticle}
\bauthor{\bsnm{Hoffmann}, \binits{N.}},
\bauthor{\bsnm{Cerqueira}, \binits{T.F.}},
\bauthor{\bsnm{Schmidt}, \binits{J.}},
\bauthor{\bsnm{Marques}, \binits{M.A.}}:
\batitle{Superconductivity in antiperovskites}.
\bjtitle{NPJ Computational Materials}
\bvolume{8}(\bissue{1}),
\bfpage{150}
(\byear{2022})
\end{barticle}
\endbibitem

\bibitem[\protect\citeauthoryear{Meredig et~al.}{2018}]{meredig2018can}
\begin{barticle}
\bauthor{\bsnm{Meredig}, \binits{B.}},
\bauthor{\bsnm{Antono}, \binits{E.}},
\bauthor{\bsnm{Church}, \binits{C.}},
\bauthor{\bsnm{Hutchinson}, \binits{M.}},
\bauthor{\bsnm{Ling}, \binits{J.}},
\bauthor{\bsnm{Paradiso}, \binits{S.}},
\bauthor{\bsnm{Blaiszik}, \binits{B.}},
\bauthor{\bsnm{Foster}, \binits{I.}},
\bauthor{\bsnm{Gibbons}, \binits{B.}},
\bauthor{\bsnm{Hattrick-Simpers}, \binits{J.}}, \betal:
\batitle{Can machine learning identify the next high-temperature
  superconductor? examining extrapolation performance for materials discovery}.
\bjtitle{Molecular Systems Design \& Engineering}
\bvolume{3}(\bissue{5}),
\bfpage{819}--\blpage{825}
(\byear{2018})
\end{barticle}
\endbibitem

\bibitem[\protect\citeauthoryear{LeCun et~al.}{2015}]{lecun2015deep}
\begin{barticle}
\bauthor{\bsnm{LeCun}, \binits{Y.}},
\bauthor{\bsnm{Bengio}, \binits{Y.}},
\bauthor{\bsnm{Hinton}, \binits{G.}}:
\batitle{Deep learning}.
\bjtitle{{N}ature}
\bvolume{521}(\bissue{7553}),
\bfpage{436}--\blpage{444}
(\byear{2015})
\end{barticle}
\endbibitem

\bibitem[\protect\citeauthoryear{David and Greental}{2014}]{david2014genetic}
\begin{bchapter}
\bauthor{\bsnm{David}, \binits{O.E.}},
\bauthor{\bsnm{Greental}, \binits{I.}}:
\bctitle{Genetic algorithms for evolving deep neural networks}.
In: \bbtitle{Proceedings of the Companion Publication of the 2014 Annual
  Conference on Genetic and Evolutionary Computation},
pp. \bfpage{1451}--\blpage{1452}
(\byear{2014})
\end{bchapter}
\endbibitem

\bibitem[\protect\citeauthoryear{Pilania
  et~al.}{2013}]{pilania2013accelerating}
\begin{barticle}
\bauthor{\bsnm{Pilania}, \binits{G.}},
\bauthor{\bsnm{Wang}, \binits{C.}},
\bauthor{\bsnm{Jiang}, \binits{X.}},
\bauthor{\bsnm{Rajasekaran}, \binits{S.}},
\bauthor{\bsnm{Ramprasad}, \binits{R.}}:
\batitle{Accelerating materials property predictions using machine learning}.
\bjtitle{Scientific reports}
\bvolume{3}(\bissue{1}),
\bfpage{2810}
(\byear{2013})
\end{barticle}
\endbibitem

\bibitem[\protect\citeauthoryear{Peterson and
  Brgoch}{2021}]{peterson2021materials}
\begin{barticle}
\bauthor{\bsnm{Peterson}, \binits{G.G.}},
\bauthor{\bsnm{Brgoch}, \binits{J.}}:
\batitle{Materials discovery through machine learning formation energy}.
\bjtitle{Journal of Physics: Energy}
\bvolume{3}(\bissue{2}),
\bfpage{022002}
(\byear{2021})
\end{barticle}
\endbibitem

\bibitem[\protect\citeauthoryear{Stanev et~al.}{2021}]{stanev2021artificial}
\begin{barticle}
\bauthor{\bsnm{Stanev}, \binits{V.}},
\bauthor{\bsnm{Choudhary}, \binits{K.}},
\bauthor{\bsnm{Kusne}, \binits{A.G.}},
\bauthor{\bsnm{Paglione}, \binits{J.}},
\bauthor{\bsnm{Takeuchi}, \binits{I.}}:
\batitle{Artificial intelligence for search and discovery of quantum
  materials}.
\bjtitle{Communications Materials}
\bvolume{2}(\bissue{1}),
\bfpage{105}
(\byear{2021})
\end{barticle}
\endbibitem

\bibitem[\protect\citeauthoryear{Dylla et~al.}{2020}]{dylla2020machine}
\begin{botherref}
\oauthor{\bsnm{Dylla}, \binits{M.T.}},
\oauthor{\bsnm{Dunn}, \binits{A.}},
\oauthor{\bsnm{Anand}, \binits{S.}},
\oauthor{\bsnm{Jain}, \binits{A.}},
\oauthor{\bsnm{Snyder}, \binits{G.J.}}:
Machine learning chemical guidelines for engineering electronic structures in
  half-heusler thermoelectric materials.
Research
(2020)
\end{botherref}
\endbibitem

\bibitem[\protect\citeauthoryear{Mazhnik and
  Oganov}{2020}]{mazhnik2020application}
\begin{botherref}
\oauthor{\bsnm{Mazhnik}, \binits{E.}},
\oauthor{\bsnm{Oganov}, \binits{A.R.}}:
Application of machine learning methods for predicting new superhard materials.
Journal of Applied Physics
\textbf{128}(7)
(2020)
\end{botherref}
\endbibitem

\bibitem[\protect\citeauthoryear{Mansouri~Tehrani
  et~al.}{2018}]{mansouri2018machine}
\begin{barticle}
\bauthor{\bsnm{Mansouri~Tehrani}, \binits{A.}},
\bauthor{\bsnm{Oliynyk}, \binits{A.O.}},
\bauthor{\bsnm{Parry}, \binits{M.}},
\bauthor{\bsnm{Rizvi}, \binits{Z.}},
\bauthor{\bsnm{Couper}, \binits{S.}},
\bauthor{\bsnm{Lin}, \binits{F.}},
\bauthor{\bsnm{Miyagi}, \binits{L.}},
\bauthor{\bsnm{Sparks}, \binits{T.D.}},
\bauthor{\bsnm{Brgoch}, \binits{J.}}:
\batitle{Machine learning directed search for ultraincompressible, superhard
  materials}.
\bjtitle{Journal of the American Chemical Society}
\bvolume{140}(\bissue{31}),
\bfpage{9844}--\blpage{9853}
(\byear{2018})
\end{barticle}
\endbibitem

\bibitem[\protect\citeauthoryear{Chakraborti}{2004}]{chakraborti2004genetic}
\begin{barticle}
\bauthor{\bsnm{Chakraborti}, \binits{N.}}:
\batitle{Genetic algorithms in materials design and processing}.
\bjtitle{International Materials Reviews}
\bvolume{49}(\bissue{3-4}),
\bfpage{246}--\blpage{260}
(\byear{2004})
\end{barticle}
\endbibitem

\bibitem[\protect\citeauthoryear{Jennings et~al.}{2019}]{jennings2019genetic}
\begin{barticle}
\bauthor{\bsnm{Jennings}, \binits{P.C.}},
\bauthor{\bsnm{Lysgaard}, \binits{S.}},
\bauthor{\bsnm{Hummelsh{\o}j}, \binits{J.S.}},
\bauthor{\bsnm{Vegge}, \binits{T.}},
\bauthor{\bsnm{Bligaard}, \binits{T.}}:
\batitle{Genetic algorithms for computational materials discovery accelerated
  by machine learning}.
\bjtitle{NPJ Computational Materials}
\bvolume{5}(\bissue{1}),
\bfpage{46}
(\byear{2019})
\end{barticle}
\endbibitem

\bibitem[\protect\citeauthoryear{Ishikawa et~al.}{2019}]{ishikawa2019materials}
\begin{barticle}
\bauthor{\bsnm{Ishikawa}, \binits{T.}},
\bauthor{\bsnm{Miyake}, \binits{T.}},
\bauthor{\bsnm{Shimizu}, \binits{K.}}:
\batitle{Materials informatics based on evolutionary algorithms: Application to
  search for superconducting hydrogen compounds}.
\bjtitle{Physical Review B}
\bvolume{100}(\bissue{17}),
\bfpage{174506}
(\byear{2019})
\end{barticle}
\endbibitem

\bibitem[\protect\citeauthoryear{Stanev et~al.}{2018}]{stanev2018machine}
\begin{barticle}
\bauthor{\bsnm{Stanev}, \binits{V.}},
\bauthor{\bsnm{Oses}, \binits{C.}},
\bauthor{\bsnm{Kusne}, \binits{A.G.}},
\bauthor{\bsnm{Rodriguez}, \binits{E.}},
\bauthor{\bsnm{Paglione}, \binits{J.}},
\bauthor{\bsnm{Curtarolo}, \binits{S.}},
\bauthor{\bsnm{Takeuchi}, \binits{I.}}:
\batitle{Machine learning modeling of superconducting critical temperature}.
\bjtitle{npj Computational Materials}
\bvolume{4}(\bissue{1}),
\bfpage{29}
(\byear{2018})
\end{barticle}
\endbibitem

\bibitem[\protect\citeauthoryear{Ward et~al.}{2016}]{ward2016general}
\begin{barticle}
\bauthor{\bsnm{Ward}, \binits{L.}},
\bauthor{\bsnm{Agrawal}, \binits{A.}},
\bauthor{\bsnm{Choudhary}, \binits{A.}},
\bauthor{\bsnm{Wolverton}, \binits{C.}}:
\batitle{A general-purpose machine learning framework for predicting properties
  of inorganic materials}.
\bjtitle{npj Computational Materials}
\bvolume{2}(\bissue{1}),
\bfpage{1}--\blpage{7}
(\byear{2016})
\end{barticle}
\endbibitem

\bibitem[\protect\citeauthoryear{Hamidieh}{2018}]{hamidieh2018data}
\begin{barticle}
\bauthor{\bsnm{Hamidieh}, \binits{K.}}:
\batitle{A data-driven statistical model for predicting the critical
  temperature of a superconductor}.
\bjtitle{Computational Materials Science}
\bvolume{154},
\bfpage{346}--\blpage{354}
(\byear{2018})
\end{barticle}
\endbibitem

\bibitem[\protect\citeauthoryear{Konno et~al.}{2021}]{konno2021deep}
\begin{barticle}
\bauthor{\bsnm{Konno}, \binits{T.}},
\bauthor{\bsnm{Kurokawa}, \binits{H.}},
\bauthor{\bsnm{Nabeshima}, \binits{F.}},
\bauthor{\bsnm{Sakishita}, \binits{Y.}},
\bauthor{\bsnm{Ogawa}, \binits{R.}},
\bauthor{\bsnm{Hosako}, \binits{I.}},
\bauthor{\bsnm{Maeda}, \binits{A.}}:
\batitle{Deep learning model for finding new superconductors}.
\bjtitle{Physical Review B}
\bvolume{103}(\bissue{1}),
\bfpage{014509}
(\byear{2021})
\end{barticle}
\endbibitem

\bibitem[\protect\citeauthoryear{{National Institute of Materials
  Science}}{2022}]{nims2011supercon}
\begin{botherref}
\oauthor{\bsnm{{National Institute of Materials Science}}}:
Super{C}on Dataset.
\url{https://mdr.nims.go.jp/collections/5712mb227}
\end{botherref}
\endbibitem

\bibitem[\protect\citeauthoryear{Hosono et~al.}{2015}]{hosono2015exploration}
\begin{barticle}
\bauthor{\bsnm{Hosono}, \binits{H.}},
\bauthor{\bsnm{Tanabe}, \binits{K.}},
\bauthor{\bsnm{Takayama-Muromachi}, \binits{E.}},
\bauthor{\bsnm{Kageyama}, \binits{H.}},
\bauthor{\bsnm{Yamanaka}, \binits{S.}},
\bauthor{\bsnm{Kumakura}, \binits{H.}},
\bauthor{\bsnm{Nohara}, \binits{M.}},
\bauthor{\bsnm{Hiramatsu}, \binits{H.}},
\bauthor{\bsnm{Fujitsu}, \binits{S.}}:
\batitle{Exploration of new superconductors and functional materials, and
  fabrication of superconducting tapes and wires of iron pnictides}.
\bjtitle{Science and Technology of Advanced Materials}
\bvolume{16}(\bissue{3}),
\bfpage{033503}
(\byear{2015})
\end{barticle}
\endbibitem

\bibitem[\protect\citeauthoryear{Gra{\v{z}}ulis et~al.}{2009}]{COD2009}
\begin{barticle}
\bauthor{\bsnm{Gra{\v{z}}ulis}, \binits{S.}},
\bauthor{\bsnm{Chateigner}, \binits{D.}},
\bauthor{\bsnm{Downs}, \binits{R.T.}},
\bauthor{\bsnm{Yokochi}, \binits{A.F.T.}},
\bauthor{\bsnm{Quir{\'{o}}s}, \binits{M.}},
\bauthor{\bsnm{Lutterotti}, \binits{L.}},
\bauthor{\bsnm{Manakova}, \binits{E.}},
\bauthor{\bsnm{Butkus}, \binits{J.}},
\bauthor{\bsnm{Moeck}, \binits{P.}},
\bauthor{\bsnm{Le~Bail}, \binits{A.}}:
\batitle{{Crystallography Open Database {--} an open-access collection of
  crystal structures}}.
\bjtitle{Journal of Applied Crystallography}
\bvolume{42}(\bissue{4}),
\bfpage{726}--\blpage{729}
(\byear{2009})
\doiurl{10.1107/S0021889809016690}
\end{barticle}
\endbibitem

\bibitem[\protect\citeauthoryear{Fujii
  et~al.}{2024}]{fujii2024efficientexplorationhightcsuperconductors}
\begin{botherref}
\oauthor{\bsnm{Fujii}, \binits{A.}},
\oauthor{\bsnm{Shimizu}, \binits{K.}},
\oauthor{\bsnm{Watanabe}, \binits{S.}}:
Efficient exploration of high-Tc superconductors by a gradient-based
  composition design
(2024).
\url{https://arxiv.org/abs/2403.13627}
\end{botherref}
\endbibitem

\bibitem[\protect\citeauthoryear{Pereti et~al.}{2023}]{pereti2023individual}
\begin{barticle}
\bauthor{\bsnm{Pereti}, \binits{C.}},
\bauthor{\bsnm{Bernot}, \binits{K.}},
\bauthor{\bsnm{Guizouarn}, \binits{T.}},
\bauthor{\bsnm{Laufek}, \binits{F.}},
\bauthor{\bsnm{Vymazalov{\'a}}, \binits{A.}},
\bauthor{\bsnm{Bindi}, \binits{L.}},
\bauthor{\bsnm{Sessoli}, \binits{R.}},
\bauthor{\bsnm{Fanelli}, \binits{D.}}:
\batitle{From individual elements to macroscopic materials: in search of new
  superconductors via machine learning}.
\bjtitle{Npj Computational Materials}
\bvolume{9}(\bissue{1}),
\bfpage{71}
(\byear{2023})
\end{barticle}
\endbibitem

\bibitem[\protect\citeauthoryear{Li et~al.}{2020}]{li2020critical}
\begin{barticle}
\bauthor{\bsnm{Li}, \binits{S.}},
\bauthor{\bsnm{Dan}, \binits{Y.}},
\bauthor{\bsnm{Li}, \binits{X.}},
\bauthor{\bsnm{Hu}, \binits{T.}},
\bauthor{\bsnm{Dong}, \binits{R.}},
\bauthor{\bsnm{Cao}, \binits{Z.}},
\bauthor{\bsnm{Hu}, \binits{J.}}:
\batitle{Critical temperature prediction of superconductors based on atomic
  vectors and deep learning}.
\bjtitle{Symmetry}
\bvolume{12}(\bissue{2}),
\bfpage{262}
(\byear{2020})
\end{barticle}
\endbibitem

\bibitem[\protect\citeauthoryear{Paszke et~al.}{2019}]{paszke2019pytorch}
\begin{botherref}
\oauthor{\bsnm{Paszke}, \binits{A.}},
\oauthor{\bsnm{Gross}, \binits{S.}},
\oauthor{\bsnm{Massa}, \binits{F.}},
\oauthor{\bsnm{Lerer}, \binits{A.}},
\oauthor{\bsnm{Bradbury}, \binits{J.}},
\oauthor{\bsnm{Chanan}, \binits{G.}},
\oauthor{\bsnm{Killeen}, \binits{T.}},
\oauthor{\bsnm{Lin}, \binits{Z.}},
\oauthor{\bsnm{Gimelshein}, \binits{N.}},
\oauthor{\bsnm{Antiga}, \binits{L.}}, et al.:
Pytorch: An imperative style, high-performance deep learning library.
Advances in neural information processing systems
\textbf{32}
(2019)
\end{botherref}
\endbibitem

\bibitem[\protect\citeauthoryear{Chai and Draxler}{2014}]{chai2014root}
\begin{barticle}
\bauthor{\bsnm{Chai}, \binits{T.}},
\bauthor{\bsnm{Draxler}, \binits{R.R.}}:
\batitle{Root mean square error ({RMSE}) or mean absolute error
  ({MAE})?--arguments against avoiding rmse in the literature}.
\bjtitle{Geoscientific model development}
\bvolume{7}(\bissue{3}),
\bfpage{1247}--\blpage{1250}
(\byear{2014})
\end{barticle}
\endbibitem

\bibitem[\protect\citeauthoryear{Brownlee}{2019}]{brownlee2019probability}
\begin{bbook}
\bauthor{\bsnm{Brownlee}, \binits{J.}}:
\bbtitle{Probability for Machine Learning: Discover How to Harness Uncertainty
  with Python}.
\bpublisher{Machine Learning Mastery}, \blocation{???}
(\byear{2019})
\end{bbook}
\endbibitem

\bibitem[\protect\citeauthoryear{Carnicom
  et~al.}{2017a}]{carnicom2017superconductivity}
\begin{barticle}
\bauthor{\bsnm{Carnicom}, \binits{E.M.}},
\bauthor{\bsnm{Xie}, \binits{W.}},
\bauthor{\bsnm{Sobczak}, \binits{Z.}},
\bauthor{\bsnm{Kong}, \binits{T.}},
\bauthor{\bsnm{Klimczuk}, \binits{T.}},
\bauthor{\bsnm{Cava}, \binits{R.J.}}:
\batitle{Superconductivity in the {N}b-{R}u-{G}e $\sigma$ phase}.
\bjtitle{Physical Review Materials}
\bvolume{1}(\bissue{7}),
\bfpage{074802}
(\byear{2017})
\end{barticle}
\endbibitem

\bibitem[\protect\citeauthoryear{Carnicom et~al.}{2017b}]{carnicom2017new}
\begin{barticle}
\bauthor{\bsnm{Carnicom}, \binits{E.M.}},
\bauthor{\bsnm{Xie}, \binits{W.}},
\bauthor{\bsnm{Klimczuk}, \binits{T.}},
\bauthor{\bsnm{Cava}, \binits{R.J.}}:
\batitle{New $\sigma$-phases in the {N}b--{X}--{G}a and {N}b--{X}--{A}l systems
  ({X= Ru, Rh, Pd, Ir, Pt, and Au})}.
\bjtitle{Dalton Transactions}
\bvolume{46}(\bissue{41}),
\bfpage{14158}--\blpage{14163}
(\byear{2017})
\end{barticle}
\endbibitem

\bibitem[\protect\citeauthoryear{Carnicom et~al.}{2018}]{carnicom2018sigma}
\begin{barticle}
\bauthor{\bsnm{Carnicom}, \binits{E.M.}},
\bauthor{\bsnm{Kong}, \binits{T.}},
\bauthor{\bsnm{Klimczuk}, \binits{T.}},
\bauthor{\bsnm{Cava}, \binits{R.}}:
\batitle{The $\sigma$-phase superconductors {N}b20. 4{R}h5. 7{G}e3. 9 and
  {N}b20. 4{R}h5. 7{S}i3. 9}.
\bjtitle{Solid State Communications}
\bvolume{284},
\bfpage{96}--\blpage{101}
(\byear{2018})
\end{barticle}
\endbibitem

\bibitem[\protect\citeauthoryear{Cook}{2023}]{cook2023data}
\begin{botherref}
\oauthor{\bsnm{Cook}, \binits{A.}}:
{Data Leakage}.
\url{https://www.kaggle.com/code/alexisbcook/data-leakage}.
[Online; accessed 9-July-2024]
(2023)
\end{botherref}
\endbibitem

\bibitem[\protect\citeauthoryear{Rodr{\'\i}guez~Pe{\~n}a and
  Garc{\'\i}a~Guerra}{2024}]{rodriguez2024periodic}
\begin{barticle}
\bauthor{\bsnm{Rodr{\'\i}guez~Pe{\~n}a}, \binits{M.}},
\bauthor{\bsnm{Garc{\'\i}a~Guerra}, \binits{J.{\'A}.}}:
\batitle{The periodic spiral of elements}.
\bjtitle{Foundations of Chemistry}
\bvolume{26}(\bissue{2}),
\bfpage{315}--\blpage{321}
(\byear{2024})
\end{barticle}
\endbibitem

\bibitem[\protect\citeauthoryear{Sommer et~al.}{2023}]{sommer20233dsc}
\begin{barticle}
\bauthor{\bsnm{Sommer}, \binits{T.}},
\bauthor{\bsnm{Willa}, \binits{R.}},
\bauthor{\bsnm{Schmalian}, \binits{J.}},
\bauthor{\bsnm{Friederich}, \binits{P.}}:
\batitle{3dsc-a dataset of superconductors including crystal structures}.
\bjtitle{Scientific Data}
\bvolume{10}(\bissue{1}),
\bfpage{816}
(\byear{2023})
\end{barticle}
\endbibitem

\bibitem[\protect\citeauthoryear{Tang et~al.}{2022}]{tang2022high}
\begin{barticle}
\bauthor{\bsnm{Tang}, \binits{F.}},
\bauthor{\bsnm{Ono}, \binits{S.}},
\bauthor{\bsnm{Wan}, \binits{X.}},
\bauthor{\bsnm{Watanabe}, \binits{H.}}:
\batitle{High-throughput investigations of topological and nodal
  superconductors}.
\bjtitle{Physical Review Letters}
\bvolume{129}(\bissue{2}),
\bfpage{027001}
(\byear{2022})
\end{barticle}
\endbibitem

\end{thebibliography}

\end{document}